\documentclass[a4paper,conference]{IEEEtran}

\usepackage{amsmath, amssymb, amsfonts,semantic,colortbl,mathrsfs,stmaryrd}
\usepackage{url}
\usepackage{subfigure}
\usepackage{algorithmic}
\usepackage{float}
\usepackage[pdftex]{graphicx}
\usepackage{caption}

\newcommand{\fc}{\mathfrak{C}}
\newtheorem{theorem}{Theorem}
\newtheorem{definition}{Definition}
\begin{document}
%
\title{Curvature-based Comparison of Two Neural Networks}

\author{\IEEEauthorblockN{Tao Yu}
\IEEEauthorblockA{Department of Mathematics\\Shanghai Jiao Tong University\\
Shanghai, China\\
Email: ydtydr@sjtu.edu.cn}
\and
\IEEEauthorblockN{Huan Long}
\IEEEauthorblockA{Department of Computer Science and Engineering\\
Shanghai Jiao Tong University\\
Shanghai, China\\
Email: longhuan@cs.sjtu.edu.cn}
\and
\IEEEauthorblockN{John E. Hopcroft}
\IEEEauthorblockA{Computer Science Department\\
Cornell Uniersity\\
New York, USA\\
Email: jeh@cs.cornell.edu}}

\maketitle

\begin{abstract}
In this paper we show the similarities and differences of two deep neural networks by comparing the manifolds composed of activation vectors in each fully connected layer of them.   The main contribution of this paper includes 1)  a new data generating algorithm which is crucial for determining the dimension of manifolds; 2) a systematic strategy to compare manifolds. Especially, we take Riemann curvature and sectional curvature as part of criterion, which can reflect the intrinsic geometric properties of manifolds. Some interesting results and phenomenon are given, which help in specifying the similarities and differences between the features extracted by two networks and demystifying the intrinsic mechanism of deep neural networks.
\end{abstract}


%
\IEEEpeerreviewmaketitle

\section{Introduction}
\label{introduction}

In recent years, neural networks have been widely used in areas including pattern recognition and machine learning. While they deliver state-of-the-art accuracy on many artificial intelligence tasks, relatively less has been known about the intrinsic mechanism of their high performance. Simply using accuracy to score neural networks neglects some fundamental information hidden behind the complicated architectures. In this paper, we compare two deep neural networks with the intension to uncover how they acquire knowledge differently. To fulfill this goal, we propose the strategy of studying the manifold structure of given neural networks.

AlexNet \cite{krizhevsky2012imagenet} is the name of a convolutional neural network, which competed in the ImageNet Large Scale Visual Recognition Challenge in 2012. As mentioned in \cite{krizhevsky2012imagenet}, AlexNet consists of five convolutional layers and three fully-connected layers with a final 1000-way softmax layer and has 60 million parameters and 650,000 neurons. We compare two neural networks AlexNet1 \cite{Alexnet1}  and AlexNet2 \cite{Alexnet2} implemented in Caffe. These two neural networks have the same structure and close accuracy but different initial parameters and were trained independently. As a metaphor, they are twins who live in the same environment and have the same education, yet they could have different learning skills and acquire different knowledge in certain aspects. 
  The work in this paper  can be easily generalized to the comparison of any two neural networks with the same inputs. In contrast to existing approaches like \cite{Raghu2017}, our framework focus on studying the intrinsic geometry of manifolds, namely, curvature information.

The remainder of this paper is organized as follows. Section \ref{manifoldandSVD} presents the geometric structure of neural network, and specifies the reason for taking curvature into account. Then a new SVD-based close data generating method is proposed, based on which we give the dimension estimation of manifolds in every fully connected layer in each AlexNet.  Section \ref{compare} mainly shows the similarities and differences of two AlexNets by calculating Riemann curvature tensor and sectional curvature tensor. The traditional 2-norm distance based work is given for comparison.  Section \ref{relatedwork} shows the related work which has been established in the last years. Finally, Section \ref{conclusion} gives our conclusions and directions for further research.

\section{Manifold and SVD-close Data} \label{manifoldandSVD}
This section contains the  mathematical conceptions  necessary for our work, along with the strategy to generate a large amount of close data, which is the  foundation for dimension estimation and local curvature calculation.

\subsection{Manifold and tangent space} \label{subsec-manifold}
A manifold $\mathcal{M}$ is a topological space that locally resembles Eulidean space near each point \cite{Wanng2011geometric}.
 The concept of manifold is central to the modern manifold leaning theory, in which data are assumed to lie  a manifold with a much lower dimension $d$ compared to that of the ambient space (the Euclidean space $\mathbb{R}^D$, $d\ll D$). The main goal is to determine $d$ and the embeding mapping $f:\mathbb{R}^d \rightarrow \mathcal{M}$ with certain local or global properties being preserved, i.e., dimension reduction. To identify $d$, we will need to determine the dimension of its tangent space. Let $\mathcal{M}\subset \mathbb{R}^{D}$ be a manifold. For every point $\textbf{p}\in \mathcal{M}$, a tangent space is a real vector space that intuitively contains the possible directions in which one can tangentially pass through $\textbf{p}$. The dimension of the tangent space at every point of a connected manifold is the same as that of the manifold itself.

 In this paper, a high dimensional data point is an activation vector formed by the output of  all neurons at some layers. For images that will be classified by AlexNet as the same label, we assume that corresponding activation vectors in a specific layer lie on the same manifold. The dimension and shape of the manifold are to be determined. For AlexNet, we will focus on three fully connected layers. As a notation,  activation vectors in the 6th (7th, 8th resp.) layer which will be recognized as label $\fc$  form a $\fc_6$-\emph{manifold} ($\fc_7$,$\fc_8$-manifold resp.), or sometimes $\fc_i$ for short. Thus we have manifolds like $Persiancat_6$, $Goldfish_8$ etc.. It follows that $\fc_6, \fc_7\subseteq \mathbb{R}^{4096}$, while $\fc_8\subseteq \mathbb{R}^{1000}$.  For each $\fc_i$-\emph{manifold}, we will need to find the dimension of the tangent space at each point.

However, with the intrinsic curse of the high dimension \cite{John2017Book}, the activation vectors of two randomly selected images will almost certainly be far from each other. As to the study of manifold, the corresponding challenge is to find many close data points around any point $\textbf{p}\in \fc_i$. They will form a local patch which can approximate the tangent space $T_{\textbf{p}}\mathcal{M}$. Traditional augmentation consists of using a combination of affine transformations to generate close images, such as cropping, rotating, flipping input images and stochastic methods \cite{Wang2017, Dyk2001}. However, previous augmentation methods aim at  increasing the accuracy and reducing overfitting. When we try to approximate the tangent space on manifolds,  one of the drawbacks of the previous methods is that they could produce data points that are not close enough to the given point $\textbf{p}$. Instead in Section \ref{subsec-closedata} we propose a new close data generating strategy which succeeds in producing a large amount of close data points on manifold efficiently.
\subsection{Riemannian manifold and  curvature} \label{subsec-Riemann}
A manifold will become a metric space when it is equipped with a metric. A Riemannian manifold $\langle \mathcal{M}, g\rangle$ is a manifold $\mathcal{M}$ endowed with an inner product $g_\textbf{p}$ at the tangent space $T_{\textbf{p}}\mathcal{M}$ at each point $\textbf{p}$ that varies smoothly from point to point in the sense that if $X$ and $Y$ are differentiable vector fields on $\mathcal{M}$, then $\textbf{p}\mapsto g_{\textbf{p}}\left(X(\textbf{p}),Y(\textbf{p})\right)$ is a smooth function \cite{Wanng2011geometric}. Three concepts closely related to the manifold-based comparison of neural network are the followings:

\begin{definition} [Riemann Curvature \cite{peterson1998}] Let $\langle \mathcal{M},g\rangle$ be a Riemannian manifold and $\nabla$ the Riemannian connection. The curvature tensor is a $(1,3)-$tensor defined by
\[\mathcal{R}(X,Y)Z=\nabla_X\nabla_YZ-\nabla_Y\nabla_XZ-\nabla_{[X,Y]}Z,\]
on vector fields $X,Y,Z$.
\end{definition}

Using Riemannian metric $g$, $\mathcal{R}(X,Y)Z$ can be changed to a $(0,4)-$tensor \cite{peterson1998}:
\[\mathcal{R}(X,Y,Z,W)=g\left(\mathcal{R}(X,Y)Z,W\right).\]
Note that this value is linear with respect to $X,Y,Z,W$, so we only need to determine this value in terms of a basis of $T_{\textbf{p}}\mathcal{M}$. 
\begin{definition}[Sectional curvature \cite{peterson1998}]
Let $\langle \mathcal{M},g\rangle$  be a Riemannian manifold, $p \in \mathcal{M}$, $u,v \in T_p\mathcal{M}$ are two linearly independent tangent vectors, the sectional
curvature of the plane $\mathbb{R}u + \mathbb{R}v$ will be defined as
\[K(u,v)=\frac{R(u,v,u,v)}{\langle u,u \rangle \langle v,v \rangle-\langle u,v \rangle^2 }.\]
where $R$ is the Riemann curvature tensor.
\end{definition}

\begin{definition}[Local Isometry \cite{lee2003smooth}] Let $\langle \mathcal{M},g\rangle$ and $\langle \mathcal{N},h\rangle $ be two Riemannian manifolds where $g$ and $h$ are Riemannian metrics on them. For a map between manifolds $F: \mathcal{M}\rightarrow \mathcal{N}$, $F$ is called local isometry if $h(dF_{\textbf{p}}(v),dF_{\textbf{p}}(v))=g(v,v)$ for all $\textbf{p}\in \mathcal{M}$, $v\in T_{\textbf{p}}\mathcal{M}$. Here $dF$ is the differential of $F$.
\end{definition}

Intrinsic geometry will be preserved under local isometric mapping, like Riemann curvature tensor and sectional curvature tensor. This is one reason for us to include  curvature tensor. Another advantage of curvatures is that they can reflect  intrinsic properties of given manifold, such as the extent to which the metric tensor is not locally isometric to that of Euclidean space, which shows the deviation of the manifold from being a flat Euclidean-space. In addition, sectional curvatures can be used to determine geometric properties of the manifold such as convexity. As for AlexNet, though the manifold of  the output layer can be  proved to be convex, the $\fc_i$ ($i=6,7,8$) manifolds could  be highly curved according to the complication of the network. In this case
the 2-norm distance based methods become less reasonable.

In brief, to have a more convincing comparison between two neural networks, we propose the strategy of comparing Riemann/sectional curvature tensor of the manifolds on the same layer of two different networks with the same input. Meanwhile the traditional Euclidean-space based method and experiments are implemented for comparative analysis. Details will be given in Section \ref{subsec-Riemann}.

\subsection{Generating close data} \label{subsec-closedata}
As we have mentioned earlier, activation vectors corresponding to random images are far from each other with high probability. Actually, according to our experiment, for the original images in ImageNet, the mean Euclidean distance between  activation vectors on AlexNet1 (similar for AlexNet2) are around  \textcolor{black}{320, 100 and 90} for $\fc_i$ ($i=6,7,8$) respectively. In order to approximate the tangent space, we need a large amount of close data around any given point $\textbf{v}$. Our trick is to treat each image as a matrix (actually three matrix corresponding to the R,G,B channels resp.), then use matrix decomposition to generate enough close data, or formally:
\begin{definition}[Original/Derivative image]
Given a real matrix $I$ of size $m\times n$, then $I=U\Sigma V^{T}$ by singular value decomposition, where $U,V$ are real orthonormal matrices of size $m\times m$ and $n\times n$ respectively, and $\Sigma$ is a $m\times n$ rectangular diagonal matrix with non-negative real singular values $\{\lambda_1,\lambda_2,\cdots,\lambda_n\}$ on the diagonal in descending order. We call the image corresponding to $I$ the original image. The image corresponding to matrix $I'$  is called derivative  image of $I$ if $I'=U\Sigma'V^{T}$, where $\Sigma'$ is a rectangular diagonal matrix with diagonal elements being $\{\alpha_1,\alpha_2,\cdots,\alpha_n\}$, where $\alpha_i\in\{0,\lambda_i\}$ for $i\in \{1,2,\cdots n\}$.
\end{definition}

Intuitively speaking, the derivative images derives from setting some of the singular values of the original matrix to zero. As the essence of SVD is discriminating noise, we are getting rid of a small part of noise from the original image. In our experiment, $m,n$ are around 500. To keep the closeness, we select the last $k_1,k_2,k_3~ (1\leq k_1,k_2,k_3 \leq 22)$ singular values of all for each R,G,B matrix of the primitive image to be zero. For each image we generate 10,648 close images. These images are inputs to both AlexNets to get the manifold. For given $\fc_i$ ($i=6,7,8$), the distance in a local patch is less than 0.9, 0.5 and 0.3 respectively, much smaller than that of the original data.

\begin{figure}[t]
\centering
\subfigure[] {\includegraphics[scale=0.15]{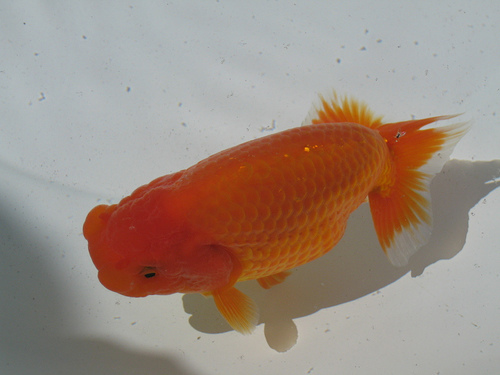}}
\subfigure[] {\includegraphics[scale=0.15]{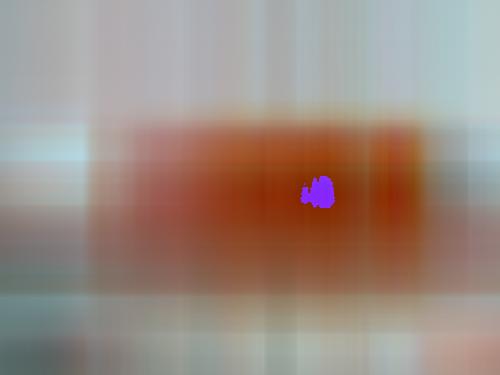}}
\subfigure[] {\includegraphics[scale=0.15]{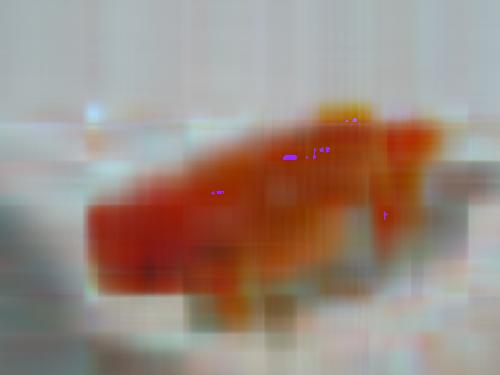}}
\caption{Images with different singular values: (a) Original $Goldfish$ image with all singular values included. (b) Derivative image of (a) with the first 3 singular values being kept. (c) Derivative image of (a) with the first 7 singular values being kept. It turns out that Alexnet classify (a) as a $Goldfish$ with probability 0.999988, (b) as a $Goldfish$ with probability 0,218679 and (c) as a $Goldfish$ with probability 0.900654. On the contrary, it is hard for human to say that (b) and (c) are of class $Goldfish$ for sure.}
\label{fig_1} \vspace{-5mm}
\end{figure}

An interesting observation in our experiment is that we find that AlexNet seems to be more  sensitive to singular values than human. As an example, for the \emph{Goldfish} image in Fig.\ref{fig_1}(a)\footnote{All images in this paper can be found in the folder n01443537 corresponding to $Goldfish$ in ImageNet.}, Alexnet only needs the first $3\sim7$ largest singular values to recognize it correctly with high probability, which is almost impossible to human. As a matter of fact, for the class of \emph{Goldfish} images, we find 288  out of the total 1236 images on which AlexNet remarkably outperforms human.

\subsection{Dimension estimation} \label{subsec-DE}
For every original image and its activation vector in the $i$-th fully connected layer ($i=6,7,8$) , we create many close images by using above strategy, and run AlexNet to get their activation vectors in $\fc_i$. As  all the vectors are close enough to form a local patch, we can now use this local patch to approximate the tangent space at given point. More precisely, to AlexNet1, for an original image $J\in \fc$, it will be mapped to a point $\textbf{v}\in\mathbb{R}^{4096}$ on the $\fc_6$-manifold. Based on $J$ we get $n$ ($n= 10,648$) derivative images $J_1,\cdots J_{n}$, whose  activation vectors in $\fc_6$-manifold will be $\textbf{v}_1^{1}, \ldots, \textbf{v}_n^1 (\textbf{v}_n=\textbf{v})$ respctively. Then we have the following data matrix which represents the local patch around $\textbf{v}$:

\begin{equation}\label{datamatrix}
X^1_{J}=\left[\textbf{v}_1^{1}, \textbf{v}_2^1, \ldots, \textbf{v}_n^1\right].
\end{equation}

\noindent It is a matrix of size $4096\times n$. We then use PCA to estimate the dimension of the $T_{\textbf{v}}\fc_6$ (also the dimension of the $\fc_6$-manifold). Similarly, we can get $X^2_{J}$ and apply the same steps to AlexNet2. We find that if we  keep the same percent of  eigenvalues of the covariance matrix, the approximate dimension of the two AlexNets will be the same. Then we repeat the work on every original images in class $\fc$. It turns out that to the images of the same class, the approximate dimensions are nearly the same.

As an example, for $Goldfish$ label, in table \ref{sample-table} we show the mean and max value of the approximate dimension, where we take the first 90\% of the sum of all eigenvalues of the covariance matrix.

\begin{table}[h]
\begin{center}
\begin{tabular}{lcccc}
\hline
Manifold & $Mean_{A1}$&$Mean_{A2}$ & $Max_{A1}$&$Max_{A2}$  \\
\hline
$Goldfish_6$  & 11.78 &  11.63 & 14 & 14 \\
$Goldfish_7$  & 10.20 &  10.02 & 13 & 13 \\
$Goldfish_8$  & 7.83 &  7.67 & 10 & 10 \\
\hline
\end{tabular}
\end{center}
\caption{By keeping the first 90\% of the eigenvalues of the covariance matrix, the mean and max values of the approximate dimension for 150 local patches of two AlexNets.}
\label{sample-table} \vspace{-3mm}
\end{table}

Note that the dimension of each layer is nearly the same for two Alexnets, also it decreases as it gets nearer to the output layer. This phenomenon dovetails nicely to the intuition that after the work of each layer, the neural network uses fewer essential features to do classification and becomes more and more certain about the attribution of the input.

\section{Comparing Two Alexnets} \label{compare}
For the two local patches in two Alexnets generated by the same image, we adopt different methods to compare them.

\subsection{Euclidean distance based method} \label{subsec-Euclidean}
Given $J$, $X^1_{J}$ and $ X^2_{J}$ have been defined in (1) (for two AlexNets). Next, we use PCA to get the dimension-reduced data matrix named as $\mathcal{DR}(X^1_J)$ ($\mathcal{DR}(X^2_J)$) from the original patch, which is of size $d\times n$ for each layer, where $d$ is the max value estimated by PCA in the table \ref{sample-table}.

 We first calculate some specific data of these two local patches on original data. For example, the mean length of each activation vector $||\textbf{v}_i||_2$, the mean distance of activation vectors in the same patch $d_{ij}$ and the mean distance of activation vectors corresponding to the same image in two patches $d(\textbf{v}_i^1,\textbf{v}_i^2)$, which are listed in TABLE \ref{dist}.

It's obvious that the mean lengths of activation vectors corresponding to the same image are similar, also the tiny difference in distance matrix for two  patches shows the closeness of the manifolds embeded in two AlexNets, which brings a question: Could it be the case that there exists a linear transformation between two AlexNets? It would be surprising if the answer is positive for the two neural networks were trained independently. To figure out the answer to this question, we proceed the work in two directions by trying to find a transformation $\mathfrak{T}$ composed of a scaling and rotation matrix $T$ and a translation vector $b$ which can minimize the root-mean-square error (RMSE) for both zero-mean normalized original data $\mathcal{N}(X_{J})$ and dimension reduced data $\mathcal{N(RD}( X_{J}))$, denoted as $L_{O-R\!M\!S\!E}$ and $L_{R\!D-R\!M\!S\!E}$:
\begin{enumerate}
  \item  $\mathcal{N}(X^1_{J})$ to $\mathcal{N}(X^2_{J})$;
  \item  $\mathcal{N(RD}( X^1_{J}))$ to $\mathcal{N(RD}( X^2_{J}))$.
\end{enumerate}
where the root-mean-square error (RMSE) is given as follows:
\[L=\sqrt{\frac{1}{n}\sum\limits_{i=1}^n(T\mathcal{N}(\textbf{v}_i^1)+b-\mathcal{N}(\textbf{v}_i^2))^{T}(T\mathcal{N}(\textbf{v}_i^1)+b-\mathcal{N}(\textbf{v}_i^2))}\]

By least square method, we can find the matrix $T$ and translation vector $b$ correspondingly. As a example, for $Goldfish$ label, the minimum error and other related data for two cases can be found in TABLE \ref{dist}.

\begin{table}[h]
\begin{center}
\begin{tabular}{r|lll}
\hline
Mean value of functions &$\fc_6$ & $\fc_7$& $\fc_8$ \\
\hline
$||\textbf{v}_i^{\tiny 1}||_2$  & 270 &  94&103 \\
$||\textbf{v}_i^{\tiny 2}||_2$  & 249 &  94&103 \\
$|d_{ij}^1-d_{ij}^2|$  & 0.0039 & 0.0001&0.0003\\
$d(\textbf{v}_i^1,\textbf{v}_i^2)$  & 350 &  122&48 \\
\hline
$||\mathcal{N}(\textbf{v}_i^{\tiny 1})||_{2}$  & 27 &  29&31 \\
$||\mathcal{N}(\textbf{v}_i^{\tiny 2})||_{2}$  & 24 &  29&31 \\
$d(\mathcal{N}(\textbf{v}_i^{\tiny 1}),\mathcal{N}(\textbf{v}_i^{\tiny 2}))$  & 36 &  42&42 \\
$r(\mathcal{N}(X^1_{J}))$ & 738 &  865&1000\\
$r(\mathfrak{T}(\mathcal{N}(X^1_{J})))$ & 618 &  860&	999 \\
$L_{O-RMSE}$ & 0.76 & 0.98 &0.92\\
\hline
$||\mathcal{N(RD}(\textbf{v}_i^{\tiny 1}))||_{2}$  & 3.70 &  3.56&3.11 \\
$||\mathcal{N(RD}(\textbf{v}_i^{\tiny 2}))||_{2}$  & 3.70 &  3.56&3.11 \\
$d(\mathcal{N(RD}(\textbf{v}_i^{\tiny 1})),\mathcal{N(RD}(\textbf{v}_i^{\tiny 2})))$  & 5.17 & 5.00&4.35 \\
$r(\mathcal{N(RD}( X^1_{J})))$ & 14 &  13&10 \\
$r(\mathfrak{T}(\mathcal{N(RD}( X^1_{J}))))$ & 14 &  13&10 \\
$L_{RD-RMSE}$ & 1.77 & 1.96 &1.93\\
\hline
\end{tabular}
\end{center}
\caption{The length $||\cdot||_2$, rank $r$, distance $d_{ij}$ and RMSE $L$ for each fully connected layer of normalized data for two cases, where $\fc=goldfish$.}\label{dist} \vspace{-3mm}
\end{table}

In the first case, the results show that (for every layer and all considered patches), $L_{O-RMSE}$ is small comparing to $d(\mathcal{N}(\textbf{v}_i^{\tiny 1}),\mathcal{N}(\textbf{v}_i^{\tiny 2}))$, yet the rank of $\mathfrak{T}(\mathcal{N}(X^1_{J}))$ will decrease a little compared to that of $\mathcal{N}(X^1_{J})$. In the second case, though the rank of $\mathfrak{T}(\mathcal{N(RD}( X^1_{J})))$ equals the rank of $\mathcal{N(RD}( X^1_{J}))$, $L_{R\!D-R\!M\!S\!E}$ between  $\mathcal{N(RD}( X^1_{J}))$ and $\mathcal{N(RD}( X^2_{J}))$ is nearly 40\% of the distance of $d(\mathcal{N(RD}(\textbf{v}_i^{\tiny 1})),\mathcal{N(RD}(\textbf{v}_i^{\tiny 2})))$. Thus, this indicates that for each point $\textbf{v}_i$, the projection of its patch to the tangent space at $\textbf{v}_i$ behaves differently separately for two Alexnets, but the patches on the manifold for two Alexnets are somehow similar.

From the experiment and analysis in this section, we can conclude that as far as manifolds are concerned, two AlexNets do behave consistently in some aspects and sharing some information, for example, similar activation vectors' length, similar distance matrix, small RMSE, similar Euclidean distance and same dimensions. On the other hand, the distinct behaviors of the projection of patches to the tangent space have revealed the existence of some differences. Yet a simple linear transformation can neither identify nor explain the difference. Moreover, the manifold itself could be highly curved. In order to compare two highly curved manifolds in high dimension and identify the similarities and differences, we take the intrinsic geometry of a manifold into consideration.

\subsection{Computation of Riemann curvature} \label{subsec-GaussEqu}
Based on the results in Section \ref{subsec-Riemann} and \ref{subsec-DE}, we can compare the Riemann curvature tensor of $\mathcal{M}$ with that of ambient space $\mathcal{\widetilde{M}}$.  Here we make use of the Gauss equation which shows the relationship between the Riemann curvature tensor of sub-manifold and that of ambient space.
\begin{theorem}[The Gauss Equation \cite{Lee1997}]
Let $\mathcal{M}$ be an $d$-dimensional embedded submanifold of a Riemannian manifold $\mathcal{\widetilde{M}}$ of dimension $D$, for any vector fields $X, Y, Z,W \in T\mathcal{M}$, the following equation holds:
\[ \begin{split}
\mathcal{\widetilde{R}}(X,Y,Z,W)=&\mathcal{R}(X,Y,Z,W)-\\
\langle \beta (X,W),\beta (Y,Z&)\rangle+\langle \beta (X,Z),\beta (Y,W) \rangle.
\end{split} \]
where $\beta (X,Y)=\widetilde{\nabla}_XY-\nabla_XY$ is the second fundamental form, $\mathcal{\widetilde{R}}$ is the Riemann curvature tensor of $\mathcal{\widetilde{M}}$ and $\mathcal{R}$ is that of $\mathcal{M}$.
\end{theorem}

Similar to the work in \cite{yangyang2017}, in our case, the Riemannian manifold $\mathcal{\widetilde{M}}$ is Euclidean space $\mathbb{R}^D$, so $\mathcal{\widetilde{R}}(X,Y,Z,W)=0$, then, the Riemann curvature of $\mathcal{M}$ is represented as
\[\mathcal{R}(X,Y,Z,W)=\langle \beta (X,W),\beta (Y,Z)\rangle-\langle \beta (X,Z),\beta (Y,W) \rangle.\]

As for the second fundamental form, we can interpret it as a measure of the difference between the Riemannian connection on $\mathcal{M}$ and $\mathcal{\widetilde{M}}$.
 Because Riemann curvature tensor is linear with respect to $X,Y,Z,W$, we only need to determine it in terms of a basis of $T_\textbf{v}\mathcal{M}$ for each point $\textbf{v}$, then we construct a local natural orthonormal coordinate frame $\{\frac{\partial}{\partial x^1},\cdots ,\frac{\partial}{\partial x^d},\frac{\partial}{\partial y^1},\cdots ,\frac{\partial}{\partial y^{D-d}}\}$ of the ambient space $\mathcal{\widetilde{M}}$ at point $\textbf{v}$, the restrictions of $\{\frac{\partial}{\partial x^1},\cdots ,\frac{\partial}{\partial x^d}\}$ to $\mathcal{M}$ form a local orthonormal frame of $T_\textbf{v}\mathcal{M}$, the last $D-d$ orthonormal coordinates $\{\frac{\partial}{\partial y^1},\cdots ,\frac{\partial}{\partial y^{D-d}}\}$ form a local orthonormal frame of $\mathcal{N}_\textbf{v}\mathcal{M}$. Then, we aim at computing $\mathcal{R}(\frac{\partial}{\partial x^i},\frac{\partial}{\partial x^l},\frac{\partial}{\partial x^j},\frac{\partial}{\partial x^k})$, denoted as $R_{iljk}$.

Under this locally natural orthonormal coordinate frame, the embeding map $f$ is redefined as \[f(x^1,x^2,\cdots,x^d)=[x^1,x^2,\cdots,x^d,f^1,f^2,\cdots,f^{D-d}].\]
where $x=[x^1,x^2,\cdots,x^d]$ are natural parameters. Then the second fundamental form is shown as
\[\beta (\frac{\partial}{\partial x^i},\frac{\partial}{\partial x^j})=\sum\limits_{\alpha=1}^{D-d}h_{ij}^{\alpha}\frac{\partial}{\partial y^{\alpha}},\]
with $h_{ij}^{\alpha},(\alpha=1,\cdots,D-d)$ being the second derivative $\frac{\partial^2}{\partial x^i\partial x^i}$ of embedding component function $f^{\alpha}$, which constitutes the Hessian matrix $H^{\alpha}=\left( \frac{\partial^2}{\partial x^i\partial x^j}\right)$, correspondingly, the Riemann curvature of $\mathcal{M}$ is represented as:
\[R_{iljk}=\sum\limits_{\alpha=1}^{D-d}(h_{ik}^{\alpha}h_{lj}^{\alpha}-h_{ij}^{\alpha}h_{lk}^{\alpha}).\]
Note that Hessian matrix is a square matrix of second-order derivatives with respect to all of the variables of scalar-values function. It can represent the concavity, convexity and the local curvature of a function. Suppose $f^{\alpha}(x^1,x^2,\cdots,x^d)$ is a multivariate function with $d$ parameters. Then the Hessian matrix $H$ of $f^{\alpha}$ is given as $H_{ij}(f^{\alpha})=\left( \frac{\partial^2}{\partial x^i\partial x^j}\right)$.  It follows that to compute the Riemann curvature of Riemannian sub-manifold $\mathcal{M}$, we just need to estimate the Hessian matrix of the embedding map $f$.

We follow the strategy in \cite{yangyang2017} to estimate the Hessian matrix. Finally we are able to calculate the Riemann curvature tensor.

\subsection{Experiments}\label{subsec-RC2FC}

\begin{figure}[h]
\centering
\subfigure[] {\includegraphics[scale=0.19]{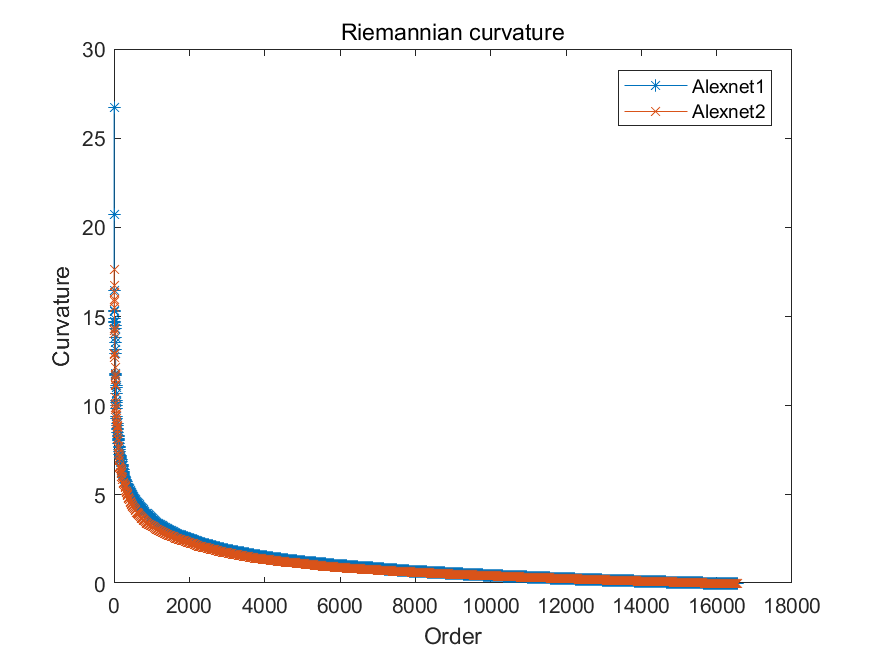}}
\subfigure[] {\includegraphics[scale=0.19]{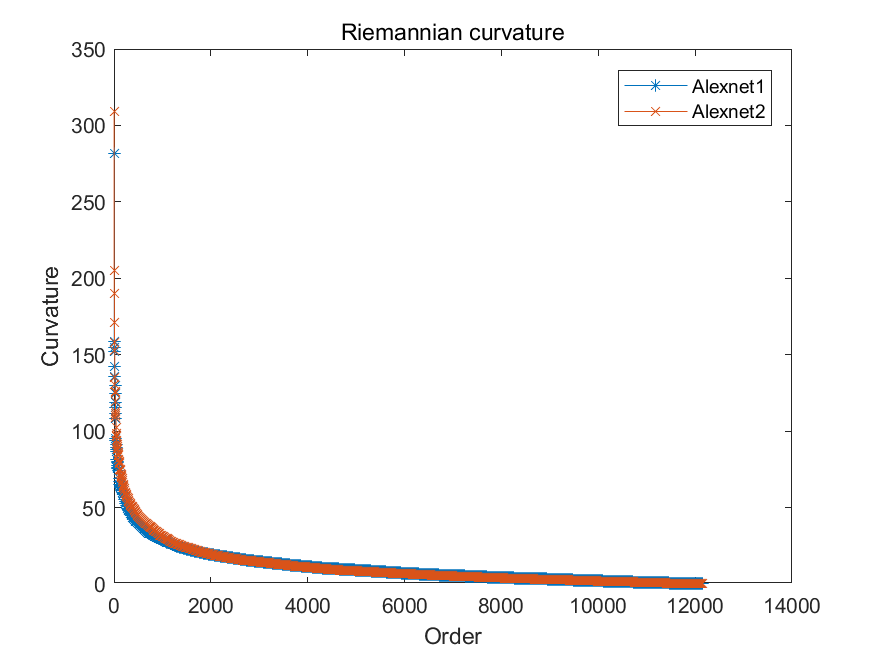}}
\subfigure[] {\includegraphics[scale=0.19]{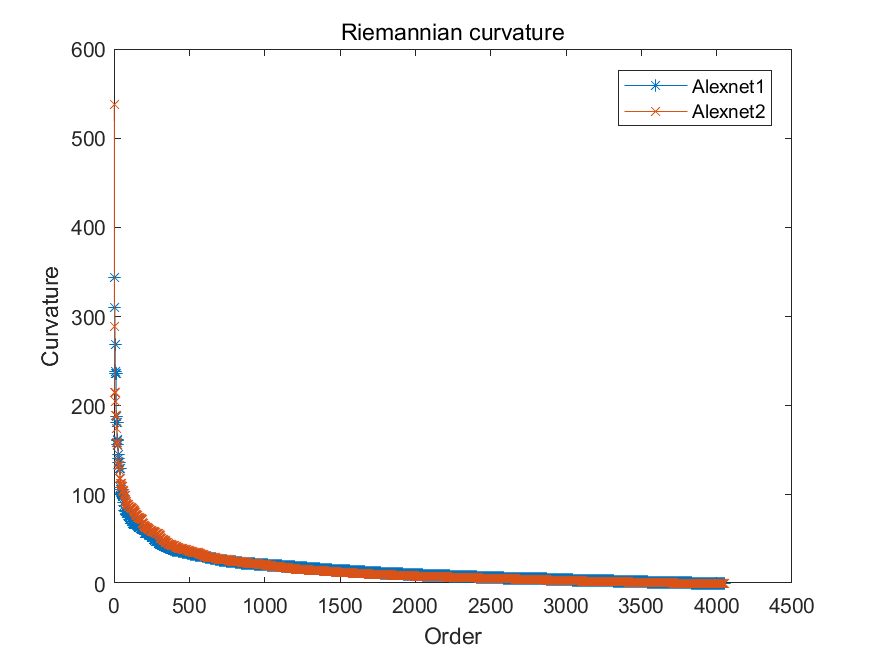}}
\subfigure[] {\includegraphics[scale=0.19]{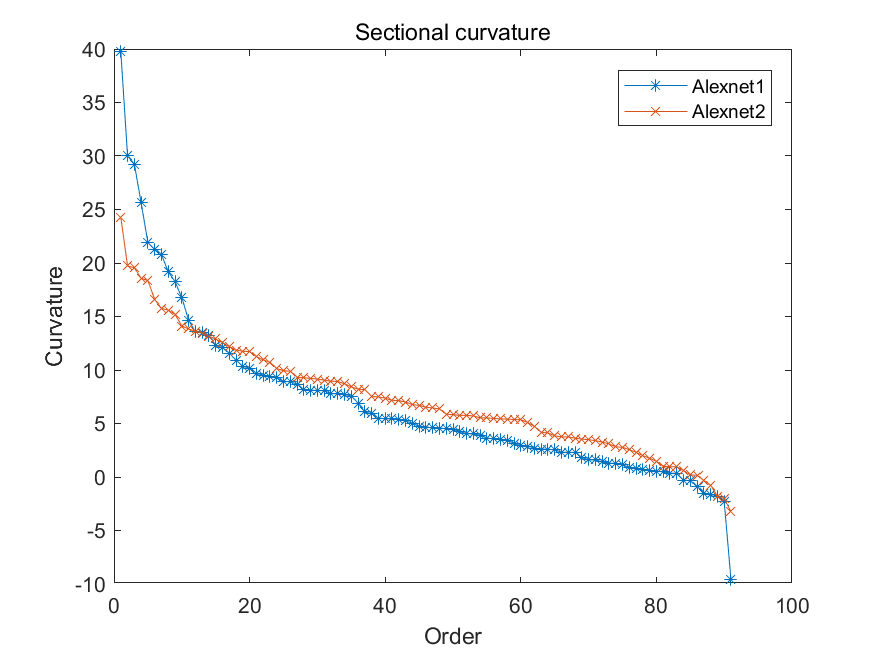}}
\subfigure[] {\includegraphics[scale=0.19]{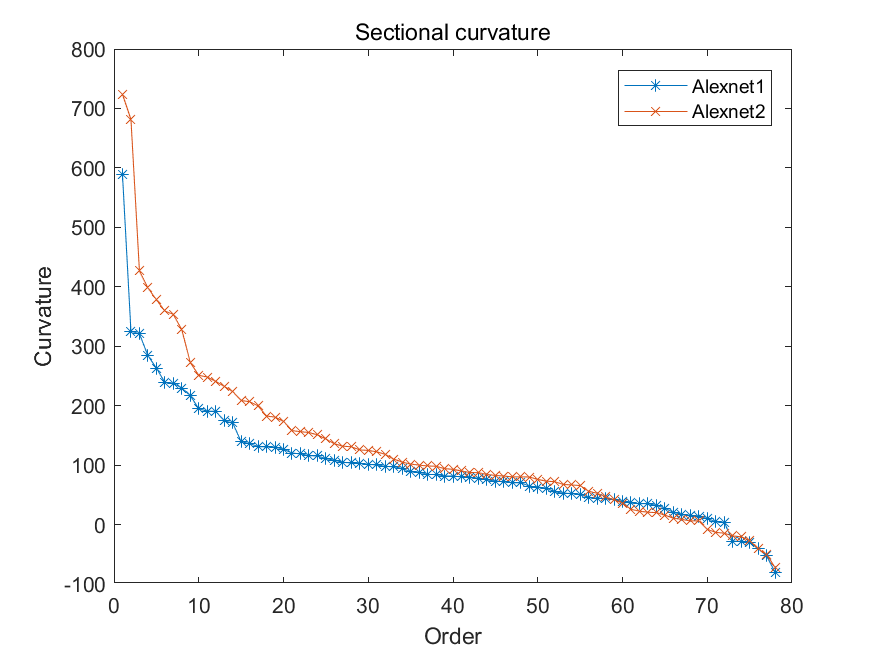}}
\subfigure[] {\includegraphics[scale=0.19]{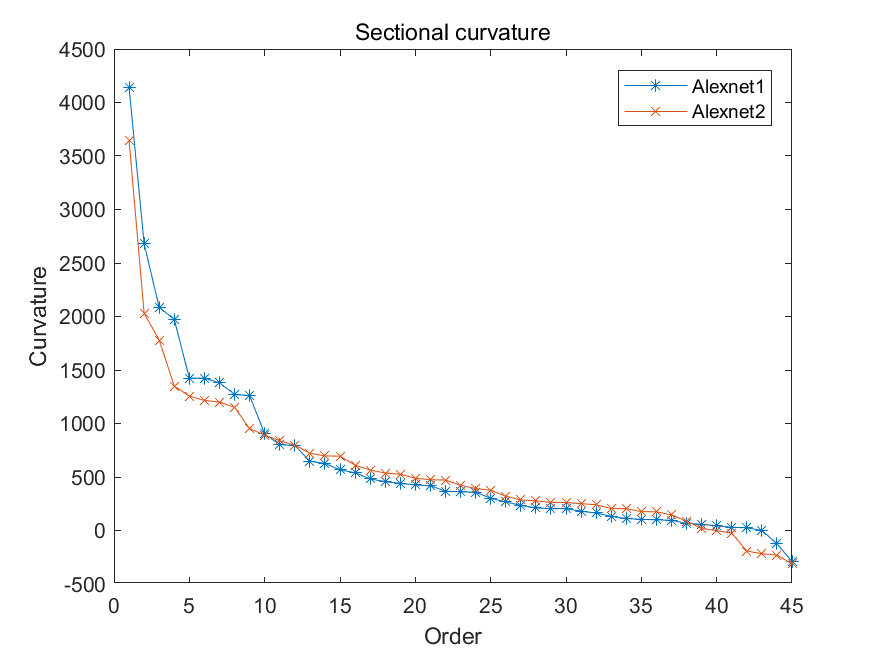}}
\caption{For a choosen $Goldfish$ image, (a), (b), (c) are the  Riemann curvature distributions of manifolds $\fc_6$  ,$\fc_7$ and $\fc_8$ respectively. (d), (e), (f) are the sectional curvature distributions of manifolds $\fc_6$  ,$\fc_7$ and $\fc_8$ respectively. It is obvious that the curvature distributions for corresponding fully connected layer almost coincide with each other for two AlexNets.}
\label{fig2}\vspace{-3mm}
\end{figure}

Applying the method mentioned above, we can get the Riemann curvature tensor of all fully connected layers.  As $R_{ijlk}=-R_{jilk}=-R_{ijkl}=R_{lkij}$,  we only calculate the positive part. At the same time, in order to do the comparison, we sort the tensor as a vector in descending order (for the correspondence of axs between two manifolds is unknown). With the computation of Riemann curvature tensor, the sectional curvatures can also be easily calculated and sorted in descending order. As an example, one result of curvature distribution curves is shown in Fig. \ref{fig2}.

\vspace{0mm}

\subsection{Analysis}\label{subsec-analysis}
\begin{figure}[h]
\centering
\subfigure[] {\includegraphics[scale=0.25]{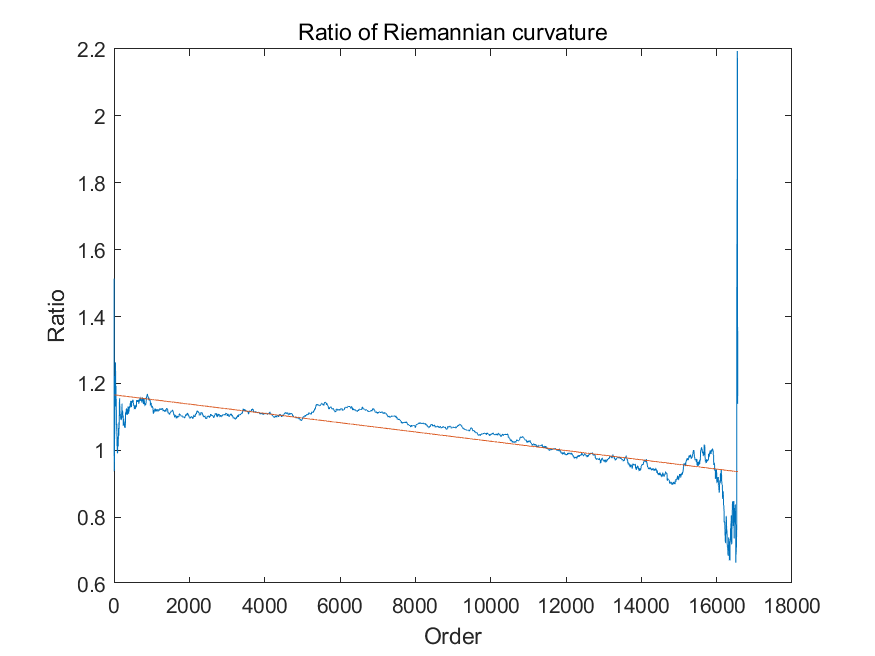}}
\subfigure[] {\includegraphics[scale=0.25]{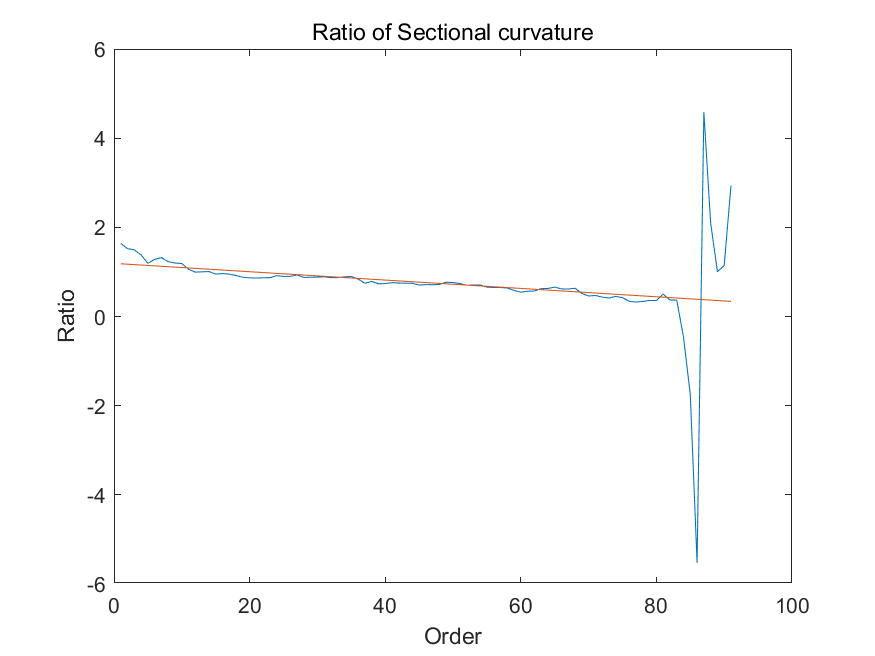}}
\caption{(a),(b) are Ratio curve of curves Fig.\ref{fig2} (a),(d). The optimum fitting straight line is painted in red. It can be seen that the similar ratio do show the similarities of two distribution.}
\label{fig3}\vspace{-3mm}
\end{figure}

In order to build a statistical result, we need to introduce a quantified characterization of the similarities of two Riemann (sectional resp.) curvature distribution curves. Here we use similar ratio. Actually, as the magnitude of the curvature ranges from $0$ to $10^3$,  we cannot just use the distance between two vectors to justify the similarity. Instead, for two curvature distribution vectors $\{a_1,a_2,\cdots,a_n\}$ and $\{b_1,b_2,\cdots,b_n\}$, we calculate the ratio vector $\{r_1,r_2,\cdots,r_n\}$ where $r_i=\frac{a_i}{b_i}$. Least square method is applied to find the optimum fitting straight line $r=ki+b$ and we define $r_0=\frac{1}{2}kn+b$ as the similar ratio of these two curvature distribution vectors. The closer $r_0$ is to 1, the more similar two curvature distributions are. An example of the ratio curve of Riemann curvature distribution, sectional curvature distribution and the optimum fitting straight line is given in Fig.\ref{fig3}.

As similar ratio reflects the similarity between images,  in TABLE \ref{S-r} we give the percentage of total images whose similar ratio lies in the specified range.

\begin{table}[h]
\begin{center}
\begin{tabular}{c|ccc}
\hline
Range &$\fc_6$ &$\fc_7$& $\fc_8$ \\
\hline
$0.8<r_{Rie}<1.2$  & 66\%&  56\%&42\% \\
$0.8<r_{Sec}<1.2$  & 30\% &  26\%&14\% \\
\hline
\end{tabular}
\end{center}
\caption{The percent of similar ratio in range $[0.8,1.2]$ of every fully connected layer, where $r_{Rie}$ ($r_{Sec}$) stands for the similar ratio for Riemann (sectional) curvature distribution and $\fc=Goldfish$.}\label{S-r} \vspace{-5mm}
\end{table}

\begin{figure}[!h]
\centering
\subfigure[] {\includegraphics[scale=0.25]{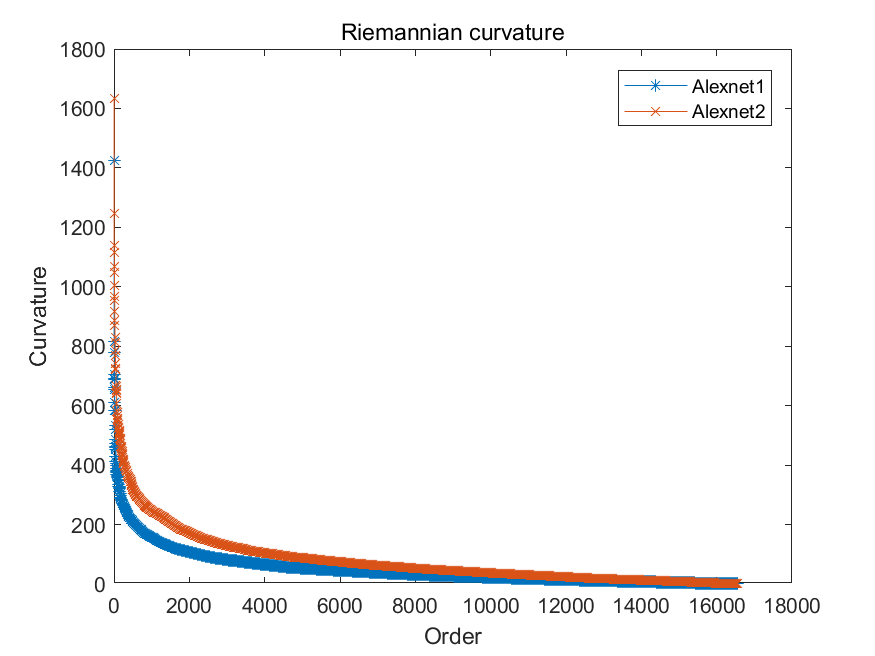}}
\subfigure[] {\includegraphics[scale=0.25]{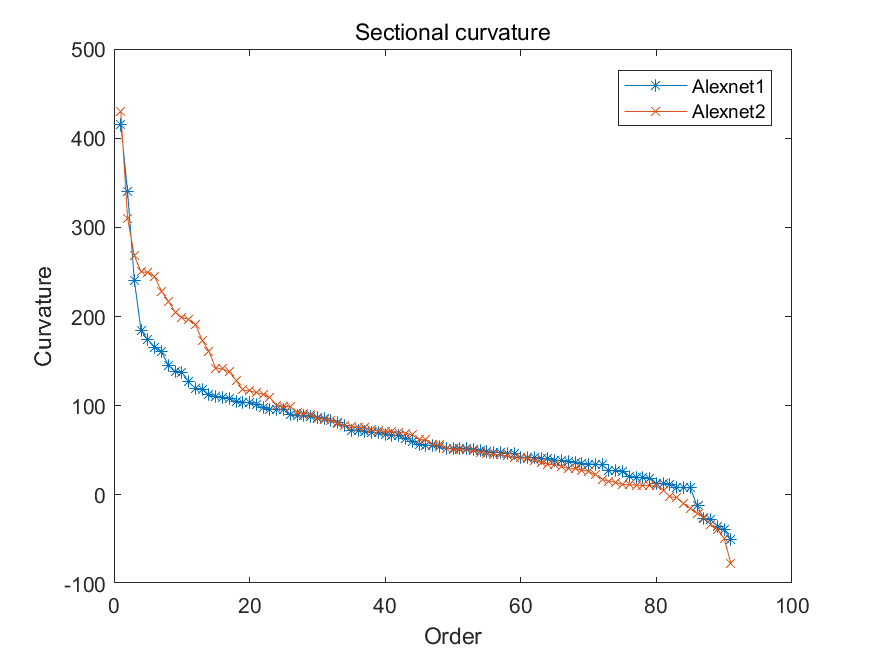}}
\caption{(a), (b) are Riemann curvature distribution and sectional curvature distribution of fully connected layers where difference is relative large.}
\label{fig4}\vspace{-3mm}
\end{figure}

The percent of similar ratio in $[0.8,1.2]$ shows that the Riemann curvature distributions and sectional curvature distributions in two Alexnets are nearly the same for a majority of images. This indicates that manifold are shaped in the same way at these points. Meanwhile, we can find that there is a correspondence between axs of the tangent space of two manifolds at corresponding points. We thus conclude that for these images, the two Alexnets are extracting the same features of the images to classify them. On the other hand, for some other images, there are pieces of Riemann curvature and sectional curvature that don't match well such as the example in Fig. \ref{fig4}. This indicates that for these images, two Alexnets are extracting different features though finally they still have the same classfication. To identify these difference is part of our future work.

\vspace{0mm}
\section{Related Work}\label{relatedwork}
The conception of curvature-aware methods have been implemented in dimension reduction. Traditional manifold learning algorithms, including LLE, LEP, LTSA, PFE \cite{Wanng2011geometric} et. al.,  always assume that the manifold is locally isometric to the Euclidean space.
However, as being highlighted in \cite{yangyang2017}, manifold $\mathcal{M}$ could be highly curved and not isometric to Euclidean space. In \cite{yangyang2017}, the authors have proposed a curvature-aware manifold learning algorithm and achieve better results. 
Comparing to our work, we include curvature tensor as a criteria for comparing neural networks.

Next, we are building a connection between manifolds and artificial neural networks. 
In \cite{brahma2016deep}, Braham et al. consider the manifold structure of the data  across the layers of a deep learning network. They have defined a set of measures which has the intrinsic assumption that the manifolds are isometric to Euclidean space. Basri et al. \cite{Basri2017} consider the ability and efficiency of deep neural networks to represent data that lies near a low-dimensional manifold in a high-dimensional space. 
Other researchers like Rifai et al. \cite{Rifai2011,Chui2016} mainly focus on representing the atlas of manifold charts. Comparatively, we use the well-known AlexNet and put no special assumption on the manifolds. Our dimension decreasing results in Section \ref{subsec-DE} could be seen as an evidence to Basri et al.'s work.

Then, for the comparison of two given neural networks, most of the known work focus on comparing the architecture and performance, for example \cite{ sze2017efficient, Liu2017,Behzad2005}.
 The work in \cite{Raghu2017,John2016} focus on comparing the representation learnt by a neuron instead of the manifold formed by the activation vectors of the same type.
Instead, what's been proposed in our work is actually the comparison of the content learnt by two deep neural networks. We are aiming at finding the features and knowledge difference between two networks. The algorithm and criteria in our work can be generalized to the comparison between any two neural networks. Especially when two neural networks have the same input and similar accuracy, the curvature-based criteria can help in telling the detail of possible different learning methods of these networks.

Finally, as for data augmentation methods, we adopt SVD-based method to generate close images. The work in \cite{Wang2017,Dyk2001} and also in \cite{krizhevsky2012imagenet}, mainly follow the idea of injecting noise into the input of a neural network during image generating.
 The intension of these methods is to enlarge the dataset using label-preserving transformations and thus does not take the closeness of activation vectors into account. The goal of the images generating in our work, however, is to create a large amount of data which is not only lable-preserving but also close to the original activation vector. Moreover, our methods can be roughly seen as removing noise and the efficiency has been shown by the experiment.

\section{Conclusion }\label{conclusion}
In this paper, two neural networks with the same structure but different parameters are compared. We aim at finding the shared knowledge, different learning methods between two networks with the help of intrinsic geometry of the manifolds embedded in them. To fulfill this goal, we have developed an effective close-data generating strategy. 
Then the dimensions of manifolds are calculated and it turns out that they are the same for two AlexNets. 

As traditional Euclidean-distance based method cannot explain the existence of both  similarity and difference between two AlexNets, our major contribution is that we take Riemann curvature distribution and sectional curvature distribution to be the criteria for the comparison of two neural networks. Our experiments show that though the two AlexNets have similar accuracy rate, their manifolds behave differently for certain inputs. 
We believe this brings more insight into the intrinsic learning behaviors of deep neural networks. It should be mentioned that the work in this paper focus on  building  similarity and difference between two networks, to figure out the specific meaning of these similarities, differences and global properties will be our next work.

The reader may have noticed that 
in this paper we only work on the manifolds for the last three fully connected layers though AlexNet has convolutional layers as well. One way is to repeat the above work for convolutional layers by taking every output of them as a long vector as in \cite{Raghu2017}.  However to our opinion, unlike the fully connected layers which is straightforward to be treated as vectors in high dimension, convolutional layer is more of matrix type, where the vector-form feature descriptor may break the geometric structure of the pixel matrix space. Redefine manifold for convolutional layers is also part of our future work.


\end{document}